\documentclass{article}
\usepackage{spconf,amsmath,graphicx,setspace}
\usepackage{amssymb}
\usepackage{algorithm}
\usepackage{algorithmic}
\usepackage{textcomp}
\usepackage{subfigure}
\usepackage{tipa}
\usepackage{cite} 
\usepackage{url}
\usepackage{float}
\usepackage{subcaption}
\usepackage{amsthm}
\usepackage{amsmath}

\makeatletter 
\renewcommand{\@thesubfigure}{\hskip\subfiglabelskip}
\makeatother
\title{GENERATIVE DATASET DISTILLATION BASED ON SELF-KNOWLEDGE DISTILLATION}
\name{Longzhen Li \quad Guang Li \quad Ren Togo \quad Keisuke Maeda \quad Takahiro Ogawa \quad Miki Haseyama \thanks{
This research was supported in part by JSPS KAKENHI Grant Numbers JP24K02942, JP23K21676, JP23K11211, JP23K11141, and JP24K23849.}
\address{Hokkaido University \\
    \{longzhen, guang, togo, maeda, ogawa, mhaseyama\}@lmd.ist.hokudai.ac.jp}}
\begin{document}
\ninept
\maketitle
%
\begin{abstract}
Dataset distillation is an effective technique for reducing the cost and complexity of model training while maintaining performance by compressing large datasets into smaller, more efficient versions. In this paper, we present a novel generative dataset distillation method that can improve the accuracy of aligning prediction logits. Our approach integrates self-knowledge distillation to achieve more precise distribution matching between the synthetic and original data, thereby capturing the overall structure and relationships within the data. To further improve the accuracy of alignment, we introduce a standardization step on the logits before performing distribution matching, ensuring consistency in the range of logits. Through extensive experiments, we demonstrate that our method outperforms existing state-of-the-art methods, resulting in superior distillation performance.
\end{abstract}
\par
\section{Introduction}
The rapid advancement of deep learning has driven the creation of increasingly large models that require vast amounts of data to achieve optimal performance~\cite{dargan2020survey}. However, this growth in data volume brings with it several significant challenges. First, storing and maintaining large datasets incurs high costs, both in terms of storage and the computational resources needed for processing~\cite{alzubaidi2021review}. Second, as datasets grow, the time required for model training becomes increasingly difficult to manage, with efficiency concerns becoming more prominent~\cite{christin2019applications}. Additionally, large-scale datasets raise important issues related to data privacy and security, particularly when handling sensitive information~\cite{abouelmehdi2018big}.
\par
To address these issues, dataset distillation has emerged as a promising solution~\cite{wang2018datasetdistillation}. This technique condenses the information from a large dataset into a smaller, more efficient version, enabling models trained on the synthetic data to achieve performance similar to those trained on the full dataset~\cite{li2022awesome}. Furthermore, dataset distillation can help alleviate privacy and sharing concerns~\cite{li2020soft, li2022compressed, li2023sharing}. As a result, dataset distillation has been applied in many downstream tasks such as continual learning~\cite{yang2023efficient}, federated learning~\cite{song2023federated}, and neural architecture search~\cite{such2020generative}. In recent years, various types of dataset distillation algorithms have been proposed, including performance matching~\cite{wang2018datasetdistillation}, gradient matching~\cite{zhao2021datasetcondensation}, trajectory matching~\cite{cazenavette2022dataset, li2024iadd}, and so on. Among these approaches, generative dataset distillation~\cite{wang2023dim} has garnered attention due to its potential to enhance flexibility and adaptability across different architectures, offering a more efficient solution for real-world applications.
\par
Generative dataset distillation aims to condense the information from large-scale datasets into a generative model rather than a static dataset~\cite{wang2023dim, li2024generative}. Unlike traditional dataset distillation methods, which produce a smaller fixed dataset, generative dataset distillation trains a model capable of generating effective synthetic data on the fly~\cite{su2024generative}. This approach has been shown to offer better cross-architecture performance compared to traditional methods, while also providing greater flexibility in the data it generates. The generative dataset distillation process typically consists of two steps. First, a generative network is trained to generate a synthetic dataset that captures the essential characteristics of the original dataset. Next, the prediction logits of synthetic and original datasets are compared. The generative network is continuously optimized based on this comparison, improving its ability to generate effective synthetic data. The accuracy of the logits matching directly influences the overall performance of the generator. However, current generative dataset distillation methods rely on relatively simple logits matching, which can constrain their effectiveness. To address this limitation, a new approach is required that enhances the precision of logits matching, enabling the generator to capture more essential information from the original dataset and improving the distillation performance.
\par
\begin{figure*}[t]
        \centering
        \includegraphics[width=15.5cm]{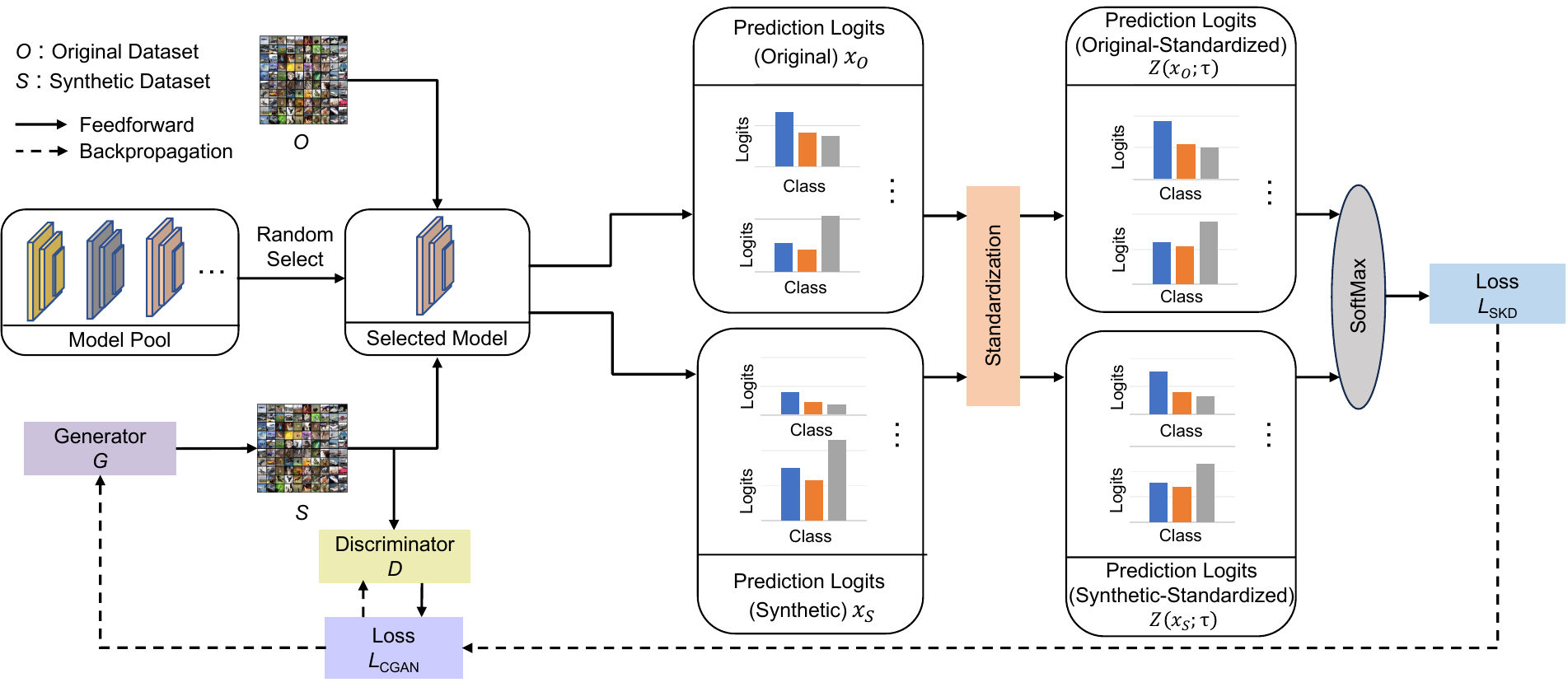}
        \caption{The distillation process of the proposed method. It involves the generator $G$ creating a synthetic dataset $S$. Both the original dataset $O$ and the synthetic dataset $S$ are then fed into a randomly selected model. The logits are standardized and their distributions are matched.}
        \label{fig1}
\end{figure*}
In this paper, we propose a novel generative dataset distillation method that leverages self-knowledge distillation to enhance the generator's performance. A key innovation of our approach is the integration of self-knowledge distillation into the optimization process, guiding the generator to more effectively align the distribution of the synthetic dataset with the original dataset. Unlike simple logits matching, distribution matching captures the overall structure and relationships in the data, leading to more robust and accurate alignment. To further improve the accuracy of alignment, we introduce a standardization step on the logits before performing distribution matching between the original and generated data. This standardization ensures consistency in the range of logits, reducing variability and improving the precision of the matching process. As a result, our method enables the generator to produce more accurate and representative synthetic data, resulting in higher-quality distillation. Extensive experiments on several benchmark datasets demonstrate the effectiveness of the proposed method compared to existing state-of-the-art methods.

Our contributions are summarized as follows.
\begin{itemize}
    \item We propose a novel generative dataset distillation method that incorporates self-knowledge distillation and improves the matching process by introducing logits standardization before performing distribution matching, which enhances the generator's ability to produce high-quality synthetic datasets.
    \item We validate the proposed method through extensive experiments, demonstrating its effectiveness and superior performance compared to existing state-of-the-art methods.
\end{itemize}
\section{Generative Dataset Distillation based on Self-knowledge Distillation}
The proposed method consists of two main steps. First, we train a generative adversarial network (GAN) to generate a synthetic dataset $S$. Next, a model is randomly selected from a model pool to align the synthetic dataset $S$ with the original dataset $O$. To improve this alignment, we incorporate self-knowledge distillation, where the distributions of the original and synthetic data are matched with the selected model. Additionally, we standardize the output logits to ensure consistency in their range and apply distribution matching to enhance the accuracy of the alignment between the synthetic and original datasets. The distillation process is illustrated in Fig.~\ref{fig1}.
\subsection{GAN Generator Training}
First, we train a conditional GAN generator to generate the synthetic dataset. The GAN framework consists of two components: a generator and a discriminator~\cite{goodfellow2014generative}. During training, the generator and the discriminator engage in a competitive process, with the generator aiming to generate increasingly realistic images, while the discriminator works to distinguish between real and generated images. This competition drives the generator to improve continuously. The training procedure for the conditional GAN is described as follows:
\begin{equation}
\begin{split}
L_{\textrm{CGAN}} & = \min_G \max_D V(D, G) \\ & = \,\, \mathbb{E}_{r \sim p(r)}[\log D(r | y)] \\
& + \,\,\, \mathbb{E}_{z \sim p(n)}[\log (1 - D(G(n | y)))],
\end{split}
\end{equation}
where $L_{\textrm{CGAN}}$ represents the conditional GAN training loss, $G$ is the generator, $D$ is the discriminator. $r$ means real data and $n$ means random noise. $y$ is the additional information that is used as input in conditional GAN. In our method, labels are utilized as this additional input to enhance the generator's ability to generate a more relevant and realistic synthetic dataset.
Once the GAN generator is trained, it can synthesize dataset $S$ as follows:
\begin{equation}
    S=G \left ( \left [ n\oplus y  \right ];\mathcal{W}  \right ),
\end{equation}
where $\oplus$ represents the concatenation operation, $n$ means random noise, $y$ is the label information, and $\mathcal{W}$ denotes the parameter of generator $G$.
\par
Unlike traditional GAN networks, which aim to generate visually realistic images, the objective of our method is to generate a synthetic dataset that effectively condenses core information from the original data. Instead of prioritizing visual fidelity, our approach focuses on preserving the essential features of the original dataset. Through iterative optimization, the generator refines the synthetic data, evolving it from random noise into a dataset that increasingly captures the core information of the original.
\par
At this stage, the dataset generated by the generator still lacks sufficient information from the original dataset. To solve this, subsequent logits matching between the synthetic and original datasets is necessary. This matching process enables continuous optimization of the generator, allowing it to progressively capture more detailed information from the original dataset in the synthetic dataset.
\subsection{Dataset Distillation via Self-knowledge Distillation}
\begin{table*}[t]
    \centering
    \small
    \caption{Comparation with data selection methods and SOTA dataset distillation methods on three benchmark datasets. The best results are highlighted in bold, while the second-best are underlined. All presented results represent the average accuracies obtained over five trials. IPC denotes images per class.}
    \label{tab1}
    \begin{tabular}{cc|ccc|ccc|ccc}
    \hline
    \multicolumn{2}{c|}{Dataset} &\multicolumn{3}{c|}{MNIST} &\multicolumn{3}{c|}{Fashion MNIST} &\multicolumn{3}{c}{CIFAR-10} \\
    \multicolumn{2}{c|}{IPC} & 1 & 10 & 50 & 1 & 10 & 50 & 1 & 10 & 50\\\hline\hline
    \multicolumn{2}{c|}{Random~\cite{zhao2021datasetcondensation}} 
    & 64.9$\pm$3.5 & 95.1$\pm$0.9 & 97.9$\pm$0.2 & 51.4$\pm$3.8 & 73.8$\pm$0.7 & 82.5$\pm$0.7 & 14.4$\pm$2.0 & 26.0$\pm$1.2 & 43.4$\pm$1.0 \\
    \multicolumn{2}{c|}{Herding~\cite{chen2010super}} 
    & 89.2$\pm$1.6 & 93.7$\pm$0.3 & 94.8$\pm$0.2 & 67.0$\pm$1.9 & 71.1$\pm$0.7 & 71.9$\pm$0.8 & 21.5$\pm$1.3 & 31.6$\pm$0.7 & 40.4$\pm$0.6 \\
    \multicolumn{2}{c|}{K-Center~\cite{chierichetti2017fair}} 
    & 89.3$\pm$1.5 & 84.4$\pm$1.7 & 97.4$\pm$0.3 & 66.9$\pm$1.8 & 54.7$\pm$1.5 & 68.3$\pm$0.8 & 21.5$\pm$1.3 & 14.7$\pm$0.9 & 27.0$\pm$1.4 \\
    \multicolumn{2}{c|}{Forgetting~\cite{toneva2019empirical}} 
    & 35.5$\pm$5.6 & 68.1$\pm$3.3 & 88.2$\pm$1.2 & 42.0$\pm$5.5 & 53.9$\pm$2.0 & 55.0$\pm$1.1 & 13.5$\pm$1.2 & 23.3$\pm$1.0 & 23.3$\pm$1.1 \\\hline\hline
    \multicolumn{2}{c|}{DC~\cite{zhao2021datasetcondensation}} 
    & 91.7$\pm$0.5 & 97.4$\pm$0.2 & 98.8$\pm$0.2 & 70.5$\pm$0.6 & 82.3$\pm$0.4 & 83.6$\pm$0.4 & 28.3$\pm$0.5 & 44.9$\pm$0.5 & 53.9$\pm$0.5 \\
    \multicolumn{2}{c|}{DSA~\cite{zhao2021differentiatble}}
    & 88.7$\pm$0.6 & 97.8$\pm$0.1 & 99.2$\pm$0.1 & 70.6$\pm$0.6 & 84.6$\pm$0.3 & 88.7$\pm$0.2 & 28.8$\pm$0.7 & 52.1$\pm$0.5 & 60.6$\pm$0.5 \\
    \multicolumn{2}{c|}{DM~\cite{zhao2023distribution}}
    & 89.9$\pm$0.8 & 97.6$\pm$0.1 & 98.6$\pm$0.1 & 71.5$\pm$0.5 & 83.6$\pm$0.2 & 88.2$\pm$0.1 & 26.5$\pm$0.4 & 48.9$\pm$0.6 & 63.0$\pm$0.4 \\
    \multicolumn{2}{c|}{CAFE~\cite{wang2022cafe}}
    & 93.1$\pm$0.3 & 97.5$\pm$0.1 & 98.9$\pm$0.2 & 73.7$\pm$0.7 & 83.0$\pm$0.4 & 88.2$\pm$0.3 & 31.6$\pm$0.8 & 50.9$\pm$0.5 & 62.3$\pm$0.4 \\
    \multicolumn{2}{c|}{KIP~\cite{nguyen2021kipimprovedresults}}
    & 90.1$\pm$0.1 & 97.5$\pm$0.0 & 98.3$\pm$0.1 & 70.6$\pm$0.6 & 84.6$\pm$0.3 & 88.7$\pm$0.2 & 49.9$\pm$0.2 & 62.7$\pm$0.3 & 68.6$\pm$0.2 \\
    \multicolumn{2}{c|}{MTT~\cite{cazenavette2022dataset}}
    & 91.4$\pm$0.9 & 97.3$\pm$0.1 & 98.5$\pm$0.1 & 75.1$\pm$0.9 & 87.2$\pm$0.3 & 88.3$\pm$0.1 & 46.3$\pm$0.8 & 65.3$\pm$0.7 & 71.6$\pm$0.2 \\
    \multicolumn{2}{c|}{FRePo~\cite{zhou2022dataset}}
    & 93.8$\pm$0.6 & 98.4$\pm$0.1 & 99.2$\pm$0.1 & 75.6$\pm$0.5 & 86.2$\pm$0.3 & 89.6$\pm$0.1 & 46.8$\pm$0.7 & 65.5$\pm$0.6 & 71.7$\pm$0.2 \\\hline\hline
    \multicolumn{2}{c|}{CGAN~\cite{mirza2014conditional}}
    & 96.1$\pm$0.7 & 97.8$\pm$0.3 & 98.4$\pm$0.3 & 81.5$\pm$0.5 & 84.0$\pm$0.2 & 86.3$\pm$0.3 & 46.4$\pm$1.2 & 62.7$\pm$0.9 & 68.1$\pm$0.8 \\
    \multicolumn{2}{c|}{DiM~\cite{wang2023dim}}
    & 96.5$\pm$0.6 & \underline{98.6$\pm$0.2} & \bfseries{99.2$\pm$0.2} & 84.5$\pm$0.4 & 88.2$\pm$0.2 & \underline{89.8$\pm$0.1} & 51.3$\pm$1.0 & 66.2$\pm$0.5 & 72.6$\pm$0.4 \\
    \multicolumn{2}{c|}{Ours (No Stand.)}
    & \underline{97.3$\pm$0.1} & 98.5$\pm$0.2 & 99.0$\pm$0.1 & \underline{85.0$\pm$0.2} & \underline{88.6$\pm$0.4} & 89.5$\pm$0.2 &  \underline{51.4$\pm$0.4} & \underline{67.3$\pm$0.5} &  \underline{73.5$\pm$0.3} \\
    \multicolumn{2}{c|}{Ours}
    & \bfseries{97.9$\pm$0.2} & \bfseries{98.7$\pm$0.1} & \underline{99.1$\pm$0.1} & \bfseries{85.5$\pm$0.4} & \bfseries{89.0$\pm$0.5} & \bfseries{89.8$\pm$0.3} & \bfseries{51.6$\pm$0.5} & \bfseries{68.2$\pm$0.4} & \bfseries{74.0$\pm$0.1} \\\hline\hline
    \multicolumn{2}{c|}{Original Dataset}
    & & 99.6$\pm$0.0 & & & 93.5$\pm$0.1 & & & 84.8$\pm$0.1 & \\\hline
    \end{tabular}
\end{table*}
\par
Before the matching process, we randomly select a model from a pool of models. Traditional dataset distillation methods typically rely on a single model for matching, which limits their ability to generalize across different architectures. In our approach, both the original and synthetic datasets are passed through the selected model to obtain prediction logits, representing the activation values just before the final output layer. By continuously matching the distribution of logits between the original and synthetic datasets, we progressively improve the GAN generator’s ability to produce synthetic data that more accurately captures the key features of the original dataset. This strategy ensures improved performance across various model architectures.
\par
Directly aligning non-standardized prediction logits can introduce some biases. For instance, variations in the logits range between the synthetic and original data can lead to instability in the matching process, potentially favoring incorrect predictions due to similar logits ranges, even if the predictions are inaccurate. To mitigate this, our method standardizes the logits before performing distribution matching. By ensuring the output logits fall within a consistent range, we achieve more accurate distribution matching, leading to improved prediction results. The detailed standardization process is shown as follows:
\begin{equation}
    Z(\boldsymbol{x}; \tau) = \frac{\boldsymbol{x} - \text{mean}(\boldsymbol{x})}{\text{std}(\boldsymbol{x}) \times \tau},
\end{equation}
where $\boldsymbol{x}$ represents the input logits vector, mean($\boldsymbol{x}$) and std($\boldsymbol{x}$) denote the mean and standard deviation of the logits respectively. The parameter $\tau$ is a reference temperature used to scale the logits during standardization. $Z(\boldsymbol{x}; \tau)$ represents the standardized logits.
\par
After performing standardization, the standardized logits are passed through the softmax function to compute the probability distribution. This allows us to calculate the probability distributions for both the original and synthetic data independently as follows:
\begin{equation}
    d(\boldsymbol{x_O}) = \text{softmax}(Z(\boldsymbol{x_O}; \tau)),
\end{equation}
\begin{equation}
    d(\boldsymbol{x_S}) = \text{softmax}(Z(\boldsymbol{x_S}; \tau)).
\end{equation}
Here, $d(\boldsymbol{x_O})$ represents the probability distribution of the original data, obtained through the softmax calculation, while $d(\boldsymbol{x_S})$ refers to the probability distribution of the synthetic data, also calculated using the softmax function. The distributions will be used for further comparison and optimization during the distillation process. 
\par
The self-knowledge distillation loss $L_{\textrm{SKD}}$ can be calculated as follows:
\begin{equation}
    L_{\textrm{SKD}} = \sum_{k=1}^{K} d(\boldsymbol{x_O})^{(k)} \log \left(\frac{d(\boldsymbol{x_O})^{(k)}}{d(\boldsymbol{x_S})^{(k)}}\right),
\end{equation}
where $K$ is the number of categories, the terms $d(\boldsymbol{x_O})^{(k)}$ and $d(\boldsymbol{x_S})^{(k)}$ denote the probability distributions of the original data and the synthetic data for the $k$th category, respectively.
Unlike the previous logits matching approach, which relied on calculating the mean squared error between the logits of the synthetic and original datasets, our method introduces a more effective solution through self-knowledge distillation. Specifically, we utilize distribution matching, which captures the overall structure and relationships within the data, resulting in a more robust and accurate alignment. To further refine this process, we incorporate a standardization step to ensure the logits remain within a consistent range, reducing variability and improving the precision of the matching process. 
\par
The total loss composed of the conditional GAN loss $L_{\textrm{CGAN}}$ and the self-knowledge distillation loss $L_{\textrm{SKD}}$ is calculated as follows:
\begin{equation}
L_{\textrm{total}} = L_{\textrm{CGAN}} + \lambda_{\textrm{SKD}} \tau^2 L_{\textrm{SKD}}.
\end{equation}
where $\lambda_{\textrm{SKD}}$ and $\tau$ denote the weight parameter and the temperature parameter, respectively. By aligning the output logits based on their distribution and standardizing them beforehand, we achieve a more precise alignment between the generated and original datasets. This approach enhances prediction accuracy and enables more effective optimization of the generator. The objective of our method is to continuously minimize the total loss, which directly optimizes the parameters $\mathcal{W}$ of the generator $G$. As the loss is reduced, the generator gradually enhances its ability to produce synthetic datasets that closely resemble the original. By minimizing the total loss, the synthetic data capture finer details and better retain the essential information from the original dataset.
\begin{figure*}[t]
        \centering
        \subfigure[MNIST]{
        \includegraphics[width=5cm]{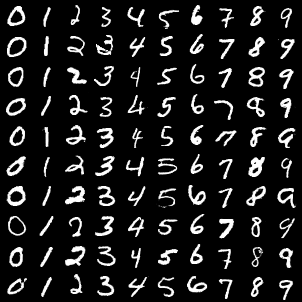}
        }
        \subfigure[Fashion MNIST]{
        \includegraphics[width=5cm]{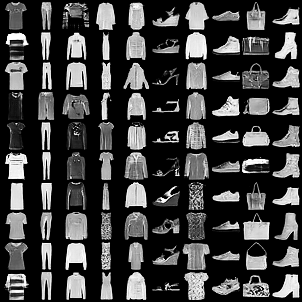}
        }
        \subfigure[CIFAR-10]{
        \includegraphics[width=5cm]{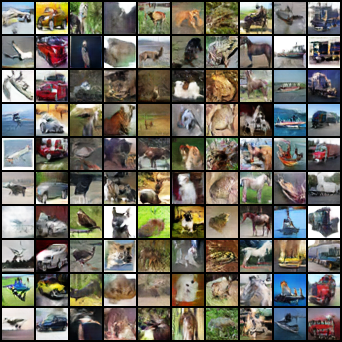}
        }
        \caption{Synthetic MNIST, Fashion MNIST, and CIFAR-10 datasets using IPC = 10.}
        \label{fig2}
\end{figure*}
\section{Experiments}
\subsection{Datasets and Comparative Methods}
We conducted extensive experiments to verify the effectiveness of the proposed method. First, we designed benchmark experiments using three datasets: MNIST~\cite{lecun1998gradient}, FashionMNIST~\cite{xiao2017fashion}, and CIFAR-10~\cite{krizhevsky2009learning}. Each dataset was categorized into 10 classes. To validate the performance of our approach, we compared it with several dataset distillation methods. The comparison includes baseline methods such as CGAN~\cite{goodfellow2014generative} and DiM~\cite{wang2023dim}, as well as four data selection methods: random selection (Random)~\cite{zhao2021datasetcondensation}, herding method (Herding)~\cite{chen2010super}, K-Center~\cite{chierichetti2017fair}, and example forgetting (Forgetting)~\cite{toneva2019empirical}. Additionally, we evaluated our method against seven state-of-the-art (SOTA) dataset distillation methods: dataset condensation (DC)~\cite{zhao2021datasetcondensation}, differentiable siamese augmentation (DSA)~\cite{zhao2021differentiatble}, distribution matching (DM)~\cite{zhao2023distribution}, aligning features (CAFE)~\cite{wang2022cafe}, kernel inducing point (KIP)~\cite{nguyen2021kipimprovedresults}, matching training trajectories (MTT)~\cite{cazenavette2022dataset}, and neural feature regression with pooling (FrePo)~\cite{zhou2022dataset}.
\subsection{Benchmark Results}
In this section, we evaluate the effectiveness of the proposed method by comparing it with other SOTA methods on three benchmark datasets. Additionally, to highlight the significance of the standardization step, we performed two sets of experiments: one using standardized prediction results (Ours) and another without standardization (Ours (No Stand.)).
Within each group, we conducted three experiments by varying the IPC (Images Per Class) and batch-size parameters. The IPC values were set to 1, 10, and 50, while batch sizes were 32, 64, and 128, respectively.
For all experiments, we utilized a model pool consisting of three neural networks, randomly selecting from ConvNet3~\cite{gidaris2018dynamic}, ResNet10~\cite{he2016deep}, and ResNet18~\cite{he2016deep}. The weight parameter $\lambda_{\textrm{SKD}}$ was set to 0.01 for FashionMNIST and CIFAR-10, and 0.001 for MNIST.
The temperature parameter $\tau$ was set to 2 in all experiments. To ensure reliability, we conducted five experiments for each IPC and batch-size configuration and averaged the results to determine the results. All experiments were conducted on an NVIDIA RTX A6000 GPU and the PyTorch framework.
\par
As shown in Table~\ref{tab1}, the proposed method demonstrates superior distillation performance across most experimental settings. In particular, for CIFAR-10, our method significantly outperforms other state-of-the-art dataset distillation methods. Furthermore, standardizing the logits leads to a notable improvement in model accuracy compared to results obtained without standardization, highlighting the effectiveness of the standardization step. The visualization results provided in Fig.~\ref{fig2} further demonstrate that our method generates more accurate and representative synthetic data.
\subsection{Cross-architecture Results}
\par
In this section, we conducted the cross-architecture experiment to demonstrate that the proposed method exhibits strong generalization performance across different network architectures. Cross-architecture performance refers to the ability to test a model trained on one architecture using a different architecture. In this experiment, we used the CIFAR-10 dataset with an IPC of 10. We evaluated the cross-architecture performance using ConvNet3~\cite{gidaris2018dynamic}, ResNet18~\cite{he2016deep}, AlexNet~\cite{krizhevsky2012imagenet}, and VGG11~\cite{simonyan2015very} as the same as previous studies~\cite{wang2023dim}.
\par
As shown in Table~\ref{tab2}, the proposed method outperforms previous dataset distillation approaches in terms of cross-architecture performance, even without standardization. After applying standardization, the results improved further. In addition to enhancing distillation performance, our method also increased the stability of the results, ensuring more robust results across different neural network architectures.
\begin{table}[t]
    \centering
    \footnotesize
    \caption{Cross-architecture comparation on CIFAR-10 dataset using IPC = 10. The best results are highlighted in bold, while the second-best are underlined. All presented results represent the average accuracies obtained over five trials.}
    \label{tab2}
    \begin{tabular}{lcccc}
    \hline
    Method & ConvNet3 & ResNet18 & AlexNet & VGG11 \\\hline\hline
    DSA~\cite{zhao2021differentiatble} & 52.1$\pm$0.4 & 42.8$\pm$1.0 & 35.9$\pm$1.3 & 43.2$\pm$0.5 \\
    KIP~\cite{nguyen2021kipimprovedresults} & 47.6$\pm$0.9 & 36.8$\pm$1.0 & 24.4$\pm$3.9 & 42.1$\pm$0.4 \\
    MTT~\cite{cazenavette2022dataset} & 64.3$\pm$0.7 & 46.4$\pm$0.6 & 34.2$\pm$2.6 & 50.3$\pm$0.8 \\
    FRePo~\cite{zhou2022dataset} & 65.5$\pm$0.4 & 57.7$\pm$0.7 & 61.9$\pm$0.7 & 59.4$\pm$0.7 \\
    DiM~\cite{wang2023dim} & 66.2$\pm$0.5 & 69.2$\pm$0.3 & 67.3$\pm$0.9 & 66.8$\pm$0.5 \\
    Ours (No stand.) & \underline{67.3$\pm$0.5} & \underline{69.6$\pm$0.5} & \underline{68.9$\pm$0.7} & \underline{68.2$\pm$0.3} \\
    Ours  & \bfseries{68.2$\pm$0.4} & \bfseries{69.9$\pm$0.3} & \bfseries{69.6$\pm$0.3} & \bfseries{69.2$\pm$0.1} \\
    \hline
    \end{tabular}
\end{table}
\section{Conclusion}
In this paper, we have proposed a novel generative dataset distillation method that incorporates self-knowledge distillation to improve the overall distillation process. A key innovation of our approach is the redesign of the logits matching process, where we employ distribution matching to better align the prediction logits between the original and synthetic datasets. To further enhance the accuracy of alignment, we apply a standardization step on logits before the matching process. This ensures a more consistent and reliable comparison between the original and synthetic datasets. Extensive experimental results show the effectiveness of the proposed method, demonstrating its superior performance in comparison to other SOTA dataset distillation methods. 
\bibliographystyle{IEEEbib}
\bibliography{refs}

\end{document}